\documentclass{article}
\usepackage[dblblindworkshop, final]{neurips_2026}
\workshoptitle{Who Verifies the Agents?}

\usepackage[utf8]{inputenc}
\usepackage[T1]{fontenc}
\usepackage{hyperref}
\usepackage{url}
\usepackage{booktabs}
\usepackage{amsfonts}
\usepackage{nicefrac}
\usepackage{microtype}
\usepackage{xcolor}
\usepackage{graphicx}
\usepackage{longtable}
\usepackage{pgfplots}
\pgfplotsset{compat=1.17}

\newcommand{\MedianFTP}{5}
\newcommand{\MedianPTP}{5}
\newcommand{\NAdmittedCorpus}{33}
\newcommand{\NBugClasses}{9}
\newcommand{\NBuggyEpisodes}{653}
\newcommand{\NDistinctPRs}{18}
\newcommand{\NEmptyFTP}{0}
\newcommand{\NFittedOnly}{2}
\newcommand{\NGoldEpisodes}{848}
\newcommand{\NLayers}{5}
\newcommand{\NManifestRows}{38}
\newcommand{\NMergedAfterCut}{8}
\newcommand{\NNoPublicFix}{14}
\newcommand{\NOracleForms}{10}
\newcommand{\NSrcTrees}{3}
\newcommand{\NSutPackages}{18}
\newcommand{\NWithUpstreamPR}{19}
\newcommand{\NWithdrawn}{5}
\newcommand{\PatchMaxFiles}{4}
\newcommand{\PatchMaxLines}{55}

\newcommand{\PatchMedianFiles}{1}
\newcommand{\PatchMedianLines}{15}

\newcommand{\PinCut}{2026-04-29}
\newcommand{\PinRelease}{1.3.12}
\newcommand{\PinSha}{6be3614}
\newcommand{\TotalCachedEpisodes}{1501}
\newcommand{\TotalFTP}{164}

\newcommand{\NControlAudible}{2}
\newcommand{\NCounterfeit}{11}

\newcommand{\NFormChecks}{162}
\newcommand{\NFormChecksHeld}{162}

\newcommand{\NPositive}{5}

\newcommand{\NAdmittedEval}{33}
\newcommand{\NUpstreamRepaired}{13}
\newcommand{\NUpstreamTotal}{57}
\newcommand{\PctUpstreamCells}{22.8}
\newcommand{\NUpstreamInstRepaired}{5}
\newcommand{\NUpstreamInst}{19}

\newcommand{\NSelfAuthoredRepaired}{17}
\newcommand{\NSelfAuthoredTotal}{42}
\newcommand{\PctSelfAuthoredCells}{40.5}
\newcommand{\NSelfAuthoredInstRepaired}{9}
\newcommand{\NSelfAuthoredInst}{14}

\newcommand{\PProvCells}{0.0773}
\newcommand{\PProvInst}{0.0397}
\newcommand{\SelfGoldAuthor}{Sonnet}
\newcommand{\NSonnetSelfRepaired}{4}
\newcommand{\PctSonnetSelf}{28.6}
\newcommand{\NSonnetUpstreamRepaired}{4}
\newcommand{\PctSonnetUpstream}{21.1}
\newcommand{\DeltaSonnetProv}{7.5}
\newcommand{\NOpusSelfRepaired}{6}
\newcommand{\PctOpusSelf}{42.9}
\newcommand{\NOpusUpstreamRepaired}{4}
\newcommand{\PctOpusUpstream}{21.1}
\newcommand{\DeltaOpusProv}{21.8}
\newcommand{\NFableSelfRepaired}{7}
\newcommand{\PctFableSelf}{50.0}
\newcommand{\NFableUpstreamRepaired}{5}
\newcommand{\PctFableUpstream}{26.3}
\newcommand{\DeltaFableProv}{23.7}
\newcommand{\NContractCompared}{11}
\newcommand{\NContractAgree}{10}
\newcommand{\NDroppedPreContract}{21}
\newcommand{\NCrashRepaired}{12}
\newcommand{\NCrashCells}{18}
\newcommand{\PctCrashRepaired}{66.7}
\newcommand{\NBehavRepaired}{18}
\newcommand{\NBehavCells}{81}
\newcommand{\PctBehavRepaired}{22.2}
\newcommand{\PCrashVsBehav}{0.0005}
\newcommand{\NCrashInstRepaired}{6}
\newcommand{\NCrashInst}{6}
\newcommand{\NBehavInstRepaired}{8}
\newcommand{\NBehavInst}{27}
\newcommand{\PCrashVsBehavInst}{0.0027}
\newcommand{\NInstUnanimous}{25}
\newcommand{\NAttempts}{1}
\newcommand{\NSeeds}{3}
\newcommand{\NCandidates}{99}

\newcommand{\NSonnetScored}{33}
\newcommand{\NSonnetRepaired}{8}
\newcommand{\PctSonnetRepaired}{24.2}

\newcommand{\NSonnetFixedButRegressed}{0}

\newcommand{\NSonnetFailedAndRegressed}{1}
\newcommand{\NSonnetRanAndFailed}{23}
\newcommand{\NSonnetBuildFailure}{1}
\newcommand{\NSonnetUnapplyable}{0}

\newcommand{\CISonnetLo}{12.8}
\newcommand{\CISonnetHi}{41.0}

\newcommand{\NOpusScored}{33}
\newcommand{\NOpusRepaired}{10}
\newcommand{\PctOpusRepaired}{30.3}

\newcommand{\NOpusFixedButRegressed}{0}

\newcommand{\NOpusFailedAndRegressed}{1}
\newcommand{\NOpusRanAndFailed}{22}
\newcommand{\NOpusBuildFailure}{0}
\newcommand{\NOpusUnapplyable}{0}

\newcommand{\CIOpusLo}{17.4}
\newcommand{\CIOpusHi}{47.3}

\newcommand{\NFableScored}{33}
\newcommand{\NFableRepaired}{12}
\newcommand{\PctFableRepaired}{36.4}

\newcommand{\NFableFixedButRegressed}{0}

\newcommand{\NFableFailedAndRegressed}{0}
\newcommand{\NFableRanAndFailed}{21}
\newcommand{\NFableBuildFailure}{0}
\newcommand{\NFableUnapplyable}{0}

\newcommand{\CIFableLo}{22.2}
\newcommand{\CIFableHi}{53.4}
\newcommand{\PModelMin}{0.21}
\newcommand{\PctModelGapWidest}{12.1}
\newcommand{\NDiscordantWidest}{6}
\newcommand{\PctPowerAtN}{19.6}
\newcommand{\NPowerWidestGap}{101}
\newcommand{\NPowerHalfGap}{410}

\newcommand{\NRuleCf}{11}
\newcommand{\NRulePos}{5}
\newcommand{\AlphaBest}{0.01}
\newcommand{\NRuleBestCaught}{4}
\newcommand{\NRuleBestFR}{0}
\newcommand{\NRuleBestResidual}{7}
\newcommand{\AlphaLoose}{0.05}
\newcommand{\NRuleLooseCaught}{7}
\newcommand{\NRuleLooseFR}{1}
\newcommand{\NRuleLooseResidual}{4}
\newcommand{\NSignalChannels}{4}

\newcommand{\NExposedRepaired}{10}
\newcommand{\NExposedCells}{33}

\newcommand{\NUnexposedRepaired}{20}
\newcommand{\NUnexposedCells}{66}

\newcommand{\PExposure}{1.0}

\newcommand{\NLOneCells}{6}

\newcommand{\NLTwoCells}{54}
\newcommand{\NLTwoRepaired}{19}
\newcommand{\NLThreeCells}{18}
\newcommand{\NLThreeRepaired}{5}
\newcommand{\NLFourCells}{15}
\newcommand{\NLFourRepaired}{6}
\newcommand{\NLSixCells}{6}

\title{RoboBugBench: Benchmarking Automated Repair of Real Robot Software\\Under an Audited Behavioral Oracle}

\author{Alexey Tuzikov\\Robouley\\Stanford University}

\begin{document}
\maketitle

\begin{abstract}
Language agents increasingly modify robot software, but whether they can \emph{repair} it is
unmeasured, because the domain supplies no oracle to inherit: a robot bug manifests as behavior, and
robot repositories ship almost no tests that capture it. We present RoboBugBench, a benchmark of
\NAdmittedCorpus{} real, execution-validated defects in a pinned ROS~2 navigation stack, in which a
candidate patch is graded by compiling it and driving the robot through a physics simulation rather
than by running a unit test. Three frontier models, each given \NAttempts{} attempt per instance,
repair \PctSonnetRepaired{}--\PctFableRepaired\%, with no pairwise difference significant; difficulty
tracks the oracle rather than the domain, at \PctCrashRepaired\% on instances graded by process
survival against \PctBehavRepaired\% on those graded by what the robot did (instance-level
$p = \PCrashVsBehavInst{}$).
Because the oracle is constructed rather than inherited we audit it adversarially, and report the
decomposition that audit yields: counterfeit patches which suppress a failure without repairing the
defect are accepted by the shipped rule \NRuleCf{}/\NRuleCf{}, while a distributional rule over the
same data rejects \NRuleBestCaught{} at no cost in false rejects. The residual is an
\emph{observability bound} --- what the observation channel discards, no predicate over it can
recover --- so discrimination must be bought with instrumentation rather than stricter criteria,
and our repair rates are bounds --- inflated by what the oracle cannot distinguish, deflated
by unanimity's rejection of correct fixes under run-to-run noise. We
release the corpus, criterion batteries, counterfeit suite and cached telemetry.
\end{abstract}

\section{Introduction}

Robot autonomy stacks are large, concurrent, safety-relevant C++ systems, and language agents are
increasingly asked to modify them. Whether an agent can \emph{repair} such a system is unmeasured,
because the instrument that makes repair measurable elsewhere does not exist in this domain.

That instrument is a test. In general-purpose software, a repair benchmark can inherit its oracle from
the repository: a maintainer wrote a test, the test encodes intent, and a patch makes it pass or
does not. Three properties come free --- the oracle \emph{exists}, it is \emph{deterministic}, and
its verdict is \emph{binary}.

Robot software gives none of the three. In ROBUST~\cite{timperley2024robust}, the field's reference
corpus of real ROS defects, only 15 of 156 runtime bugs ship a test that exercises them at all, and
its authors pitch the 115 system-level bugs --- whose detection ``relies heavily on human oracles''
--- as an open agenda.\footnote{ROBUST catalogues ROS~1 systems from 2010--2018 while our substrate is
ROS~2, so we import this premise across a version boundary. We know of no equivalent ROS~2 census; our
own corpus is consistent with it, and in \S\ref{sec:oracle} the one upstream test we could borrow
fails to discriminate its own bug.} Behavior lives in a continuous control loop coupled to a physics
simulator: a run is stochastic across repetitions, the ``result'' is a trajectory rather than an
assertion,
and correctness is an envelope of acceptable behaviors rather than a single trace
\cite{afzal2022mithra}. A benchmark for this domain must therefore \emph{construct} its oracle --- and
then carries an obligation a test-inheriting benchmark never has: showing that the oracle it built is
right.

RoboBugBench takes on both. It presents an agent with a real defect in a pinned ROS~2 navigation
stack and the source files that contain it, and grades the returned patch by compiling it into the
stack and running the robot through a Gazebo~\cite{koenig2004gazebo} mission under a criterion
battery calibrated against cached buggy and reference behavior --- then subjects that oracle to a
two-sided audit and an adversarial counterfeit suite and reports what the audit found.

Two documented failure modes of execution-based repair benchmarks shaped that construction from the
outset: their oracles accept incorrect patches at high rates
\cite{yu2025utboost,li2026probe,wang2025solved}, and their instances leak into training data, so a
measured score partly reflects memorization rather than reasoning \cite{liang2026illusion}. We take
both up in \S\ref{sec:related}; we designed \emph{against} them, which is why contamination dating
and two-sided oracle measurement appear below as contributions rather than caveats.

\paragraph{Contributions.}
\begin{enumerate}
\item \textbf{An execution-graded repair benchmark for real robot software.}
\NAdmittedCorpus{} defects across \NSutPackages{} packages and \NSrcTrees{} source trees, each
admitted only after its buggy and reference behaviors were shown to separate under simulation
(\TotalCachedEpisodes{} cached episodes), each carrying a resolvable reference patch (median
\PatchMedianLines{} changed lines), and each organized by an induced mechanism taxonomy
(\NLayers{} architectural layers, \NBugClasses{} mechanism classes, \NOracleForms{} oracle forms) so
results are diagnosable rather than a single scalar.
\item \textbf{An outcome taxonomy that separates repair from collateral damage.} A ten-value taxonomy
replaces the repaired/not-repaired binary, distinguishing a patch that fixes the defect from one that
fixes it \emph{by breaking behavior that previously worked} --- the failure mode a behavioral
benchmark is uniquely exposed to, and which share-fixed alone cannot see.
\item \textbf{A three-model evaluation with a fully disclosed protocol.} Three frontier models,
\NAttempts{} attempt per instance, \NSeeds{} replicate episodes per evaluation, prompt reproduced
verbatim (Appendix~\ref{app:prompt}). Difficulty is concentrated exactly where the oracle becomes
behavioral (instance-level $p = \PCrashVsBehavInst{}$), and the two deepest-logic layers are unsolved
by every model.
\item \textbf{A corpus dated so that contamination cannot explain the results.} The goal is to make
memorization an implausible account of any score, checkable rather than asserted: \NMergedAfterCut{}
upstream fixes merged strictly \emph{after} the \PinCut{} pin cut, and \NNoPublicFix{} reference fixes
exist in no public record at all, so a model's training-cutoff date alone shows most of the corpus
could not have been seen.
\item \textbf{An audited oracle, and a decomposition of where its discrimination comes from.} We
measure both error directions against blind labels and attack the oracle with purpose-built
counterfeit patches on \NRuleCf{} instances. Sweeping decision rules over the same episodes separates
the acceptance failure into a fixable part (aggregation: \NRuleBestCaught{}/\NRuleCf{} recovered at
zero false-reject cost) and an irreducible part (the observability bound). The result is a measured
detection/false-reject trade-off curve for behavioral oracles, not a bare negative.
\end{enumerate}

\section{Related work}
\label{sec:related}

\paragraph{Execution-based repair benchmarks.} SWE-bench \cite{jimenez2024swebench} established the
template we depart from: mine repository history for issue-resolving pull requests, keep those
contributing tests that fail before and pass after, and score a patch by running those tests. Its
follow-up literature catalogues the ways an execution oracle can be wrong. SWE-bench Verified \cite{openai2024verified} audited
1{,}699 instances with 93 annotators and kept 500, flagging 61.1\% for unfair tests --- but measured
only whether tests wrongly \emph{reject} a valid patch. Third
parties supplied the opposite direction and found it large: 345 wrongly-passed patches
\cite{yu2025utboost}; 77\% of Verified instances admitting at least one incorrect passing patch
\cite{li2026probe}; 29.6\% of ``resolved'' patches behaviorally divergent from the reference
\cite{wang2025solved} --- a critique our design answers by construction, since our grading \emph{is}
behavioral rather than a proxy for it. Memorization is a second live confound: models recover up to
76\% of buggy file paths from issue text alone \cite{liang2026illusion}, and 28.5\% of tasks admit a
shortcut worth $+14.14$\,pp \cite{rajan2026hackability}. DeepSWE \cite{huang2026deepswe} responds by
building original tasks with never-published solutions, trading realism for contamination-proofing.
Each of these findings has a design consequence here. One-sided auditing we reject outright: we measure
both error directions and publish the pair or neither (\S\ref{sec:oracle}). Incorrect-patch acceptance
we do not escape --- we run the critique's own probe on ourselves and do \emph{worse}, which
\S\ref{sec:bound} root-causes rather than hides. Contamination we answer by construction, with
\NMergedAfterCut{} post-cut fixes, \NNoPublicFix{} never published, and no issue prose in the prompt.
Shortcut solutions we answer with an explicit P2P battery (\S\ref{sec:bench}). The scale critique we
inherit more acutely than any of them: \NAdmittedCorpus{} instances over \NSrcTrees{} source trees,
against 2{,}294 over 12 repositories. Classical APR established the underlying
phenomenon --- patches can pass tests without being correct, typically by deleting functionality
\cite{qi2015analysis,smith2015cure}.

\paragraph{Verifier auditing.} Measuring a verifier in both error directions is established practice
that we adopt rather than introduce. RLVR verifier fuzzing \cite{ray2026fuzzing} generates adversarial
completions against a stricter reference verifier and reports false-positive and false-negative rates
directly; DeepSWE audits verifier quality through judge disagreement \cite{huang2026deepswe}. New
here is the setting: a physics-simulation verifier changes both what an adversarial example looks
like and what the audit can conclude (\S\ref{sec:bound}).

\paragraph{Robotics testing and oracles.} Deciding whether observed behavior is correct is the
classical oracle problem \cite{barr2015oracle}, and measuring a deployed oracle's own error rates is
an established task \cite{jahangirova2016oracle} that \S\ref{sec:bound} carries out with behavioral
counterfeits in place of mutants. ROBUST \cite{timperley2024robust} catalogues 221 real ROS bugs and
supplies our test-scarcity premise; misconfiguration studies \cite{canelas2024misconfig} map the
failure surface our largest layer (runtime configuration) draws from; Afzal et al.\
\cite{afzal2020challenges} name absence of an oracle and distrust of simulation among nine
practitioner challenges. RoboFuzz \cite{kim2022robofuzz} fuzzes ROS systems against semantics-aware
oracles to \emph{find} correctness bugs; Mithra \cite{afzal2022mithra} learns cyber-physical oracles
without per-instance ground truth; Jahangirova et al.\ \cite{jahangirova2021metrics} evaluate
candidate oracle metrics for autonomous driving; simulation flakiness
\cite{amini2024flaky,chance2022determinism} motivates our cross-replicate aggregation. Closest in
machinery is CPS falsification: signal temporal logic and its robust semantics grade a trace by a
continuous margin, not a binary verdict \cite{maler2004monitoring,fainekos2009robustness}; Breach
\cite{donze2010breach} and S-TaLiRo \cite{annpureddy2011staliro} minimize that margin over inputs
to find violating traces \cite{bartocci2018specification,corso2021blackbox}; VerifAI
and Scenic add scenario-level environment search for AI-based systems
\cite{dreossi2019verifai,fremont2019scenic}. We adopt this machinery rather than invent it --- the
release ships an offline rtamt-based robustness monitor \cite{nickovic2020rtamt}, its margins as
severity metadata, not verdicts; what \S\ref{sec:bound} adds is a measured detection/false-reject
comparison of decision semantics on real repair candidates. None grades a repair agent against
real robot defects by executing the repaired system, to our knowledge.

\section{RoboBugBench}
\label{sec:bench}

\subsection{Task formulation}

An instance is a defect in a specific package of a pinned robot software stack, together with
everything needed to reproduce and grade it: a \emph{mission} (a navigation goal in a Gazebo world), a
\emph{fault trigger} that exercises the defect, a \emph{control mission} that exercises the same code
path legitimately, a \emph{criterion battery}, and a resolvable \emph{reference patch}. The stack as
shipped is the buggy version; the reference side runs a fix overlay.

A solver receives a harness-authored context: a task statement, the observed failure of the buggy run,
and the complete current text of every source file the reference fix touches. It receives no issue
title or body, commit message, upstream link, repository history, or test. It returns a
unified diff. The full prompt is reproduced verbatim in Appendix~\ref{app:prompt}.

Grading compiles the returned patch into the stack and executes the mission. This is the load-bearing
difference from a test-graded benchmark: the verdict is a property of the running robot, so a patch
that looks correct statically but degrades control behavior is scored as what it is.

\subsection{Corpus construction}

Every instance is cut against one pin: Nav2~\cite{macenski2020marathon} \PinRelease{} at \texttt{\PinSha{}}, cut \PinCut{}.
Candidates are mined from upstream history, issue reports and code audit, then authored into runnable
packages and put through an admission gate that requires the buggy and reference arms to separate
under repeated simulation. \NAdmittedCorpus{} of \NManifestRows{} authored instances survive;
\NWithdrawn{} were withdrawn under a pre-registered falsifier battery, including cases where the
apparent discriminator turned out to be an artifact of mission timing rather than the defect.
Containers run with no network.

\paragraph{Three provenance modes.} \emph{Latent}: the upstream fix postdates our pin, so the defect
is in our source today. \emph{Resurrected}: a pre-pin fix reverse-applied. \emph{Self-fixed}: the
defect is real and present in pinned code but no upstream fix exists, so the reference fix is ours ---
the last two being inexpressible by pull-request mining. \NWithUpstreamPR{} instances trace to a real
upstream PR (\NDistinctPRs{} distinct), \NMergedAfterCut{} of those fixes merged strictly after the pin
cut, and \NNoPublicFix{} exist in no public record at all.

\subsection{The behavioral oracle}

A criterion battery is drawn from \NOracleForms{} criterion forms over telemetry --- distance to goal,
collision, velocity envelope, process liveness, action-server state, planner outcome --- and split
into fail-to-pass (F2P) criteria, which the buggy arm fails and a correct patch must satisfy, and
pass-to-pass (P2P) criteria, which the buggy arm already satisfies and a patch must not break. The
corpus carries a median of \MedianFTP{} F2P and \MedianPTP{} P2P criteria per instance
(\TotalFTP{} F2P total); \NEmptyFTP{} instances have an empty F2P set.

Two design decisions distinguish this from a unit-test battery. First, \emph{verdicts aggregate across
replicate episodes}. A single Gazebo episode is stochastic; each evaluation runs \NSeeds{} replicate
episodes per mission and the criterion verdict is taken over the set. Second, \emph{the P2P battery is
anti-gaming machinery, not bookkeeping}. In a behavioral benchmark the cheapest way to satisfy a
criterion is often to stop the robot: a zero-velocity ``fix'' trivially satisfies \emph{no collision}.
Where a test-graded benchmark can leave non-regression implicit in an existing suite, we must make it
an explicit semantics, and the control mission exists so that P2P has legitimate behavior to protect.
Thresholds are withheld from solvers; the criterion \emph{forms} are disclosed.

\subsection{Outcome taxonomy}

Reporting share-fixed alone conflates a solver that produced nothing with one that produced a patch
that made the system worse. We score every attempt into one of ten outcomes:
\texttt{no\_submission}, \texttt{empty\_patch}, \texttt{invalid\_submission}, \texttt{unapplyable},
\texttt{build\_failure}, \texttt{ran\_and\_failed}, \texttt{fixed\_but\_regressed},
\texttt{failed\_and\_regressed}, \texttt{repaired}, \texttt{harness\_error}.

Two choices are deliberate. \texttt{harness\_error} is excluded from rates but always reported with a
count, so infrastructure failure can never silently shrink a denominator. And the regression outcome
is \emph{split}: a patch that fixes the defect while breaking previously-passing behavior
(\texttt{fixed\_but\_regressed}) is a different claim about a solver than one that does neither
(\texttt{failed\_and\_regressed}). Pooling them would let a merely bad patch inflate a headline that
reads as the anti-gaming result.

\subsection{Corpus composition}

\begin{table}[t]
\caption{The corpus by architectural layer, defect mechanism and oracle shape (33 instances). Concentration is real and reported: L2\_runtime\_config accounts for 18 of 33 (55\%) and input\_validation for 17 of 33 (52\%).}
\label{tab:taxonomy}
\centering\small
\begin{tabular}{@{}lr@{\hspace{2em}}lr@{\hspace{2em}}lr@{}}
\toprule
\textbf{Layer} & \textbf{n} & \textbf{Mechanism} & \textbf{n} & \textbf{Oracle shape} & \textbf{n} \\
\midrule
L2\_runtime\_config & 18 & input\_validation & 17 & behavioral & 20 \\
L3\_planning & 6 & dynamic\_reconfigure & 7 & crash\_survival & 6 \\
L4\_perception & 5 & memory\_safety & 3 & behavioral\_abort & 2 \\
L1\_algorithm & 2 & behavioral\_degradation & 1 & action\_stall & 1 \\
L6\_bt\_engine & 2 & index\_based\_progress\_tracking & 1 & behavioral\_goal\_abort & 1 \\
 &  & iterator\_underflow & 1 & behavioral\_position & 1 \\
\midrule
\multicolumn{6}{@{}l@{}}{\footnotesize Provenance: latent 29, resurrected 2, self\_fixed 2} \\
\bottomrule
\end{tabular}
\end{table}

The corpus spans \NSrcTrees{} source trees, \NSutPackages{} packages, \NLayers{} architectural layers
and \NBugClasses{} mechanism classes; Table~\ref{tab:taxonomy} gives the cross-tabulation. The layers
are algorithm internals (\texttt{L1}), runtime configuration and parameter handling (\texttt{L2}),
planning (\texttt{L3}), perception and costmap construction (\texttt{L4}), and the behavior-tree
engine (\texttt{L6}); the numbering is inherited from a wider authoring taxonomy (one layer has no
admitted instance); we keep the labels rather than renumber and break the released metadata. Instances
are referred to by mechanism throughout, with per-instance identifiers confined to the released
metadata and Appendix~\ref{app:corpus}. Reference fixes are small and surgical --- median \PatchMedianFiles{} file and \PatchMedianLines{}
changed lines, at most \PatchMaxFiles{} files and \PatchMaxLines{} lines --- so difficulty lies in
localization and mechanism, not volume of code. Behind each instance sit \NBuggyEpisodes{} valid buggy and \NGoldEpisodes{} valid
reference episodes.

\paragraph{Coverage we did not achieve.} The corpus is concentrated. Two taxonomy categories we
explicitly targeted --- geometry and collision-checking gaps for non-trivial robot footprints, and one
perception-layer class --- remain at or near zero after dedicated harvesting rounds, and
\NFittedOnly{} instances are graded by distribution-fitted bands rather than an authored semantic
discriminator. Both are labelled in the released metadata rather than smoothed away.

\section{Oracle validation}
\label{sec:oracle}

A constructed oracle is a research artifact in its own right and we treat it as one: we measure it in
both error directions against labels assigned independently of it, and we attack it. On the
two-sided measurement, dual-reference recalibration brought the battery to \textbf{1 false reject in
6} at an unchanged \textbf{2 false accepts in 8} on a like-for-like comparison, down from an initial
43.8\% false-reject rate that was a grader defect, not a finding about robots. We give
fractions rather than percentages deliberately: at these denominators a percentage implies precision
the measurement does not have; an earlier, more flattering pairing was retracted as not
like-for-like. Three further failure modes --- a liveness
criterion a stationary robot satisfies, a validation check mathematically incapable of failing, and an
instrument that degraded to a constant while passing its own gates --- were each found on our own
harness and are documented in Appendix~\ref{app:oracle}; the first is the origin of the P2P battery of
\S\ref{sec:bench}, and the others produced the invariant that \emph{a check which can degrade into a
constant is worse than no check}.

\subsection{Adversarial audit: how much can a behavioral oracle discriminate?}
\label{sec:bound}

The standard remedy for oracle inadequacy is to add the missing test. We tested whether that remedy is
available in a behavioral setting, and found that it is bounded in a way worth stating precisely.

\paragraph{Setup.} We authored one counterfeit patch for each of \NCounterfeit{} instances --- the
corpus as it stood when the audit ran, a subset of the \NAdmittedCorpus{} released here --- from a
fixed fault-operator set: read-only-parameter, blanket-no-op-callback, overfit-to-fault-value,
over-broad-rejection, partial-guard. Each suppresses the observed failure without repairing the defect
--- the functionality-deleting shape known from classical APR \cite{qi2015analysis,smith2015cure}. As a
positive control we authored \NPositive{} held-out correct fixes distinct from the reference. The suite
has not been extended to instances added since, so every number in this subsection is on
\NCounterfeit{} instances, not \NAdmittedCorpus{}.

\paragraph{Result under the shipped rule.} All \NCounterfeit{} counterfeits score \texttt{repaired}
and \NFormChecksHeld{} of \NFormChecks{} criterion checks hold across both missions and every
episode;
all \NPositive{} held-out positives are accepted --- so the positive column alone cannot distinguish a
calibrated oracle from one that accepts everything, which is exactly why the negative column had to be
run.

\paragraph{This is not a patch-application artifact.} Three earlier false discoveries in this project
were exactly that, so we verified per instance that the counterfeit source is in the overlay, the
shared object was compiled after the patch landed, the trigger was delivered, and the unpatched arm
genuinely fails. The chain holds for all \NCounterfeit{}.

\paragraph{How much of this is the rule, and how much is the channel?} Acceptance of
\NCounterfeit{}/\NCounterfeit{} alone would be uninformative: it could mean the
information is absent from the observations, or merely that our aggregation asks the wrong question.
We separated the two by re-scoring the same episodes under a different decision rule. The shipped rule
asks, per seed, \emph{does this episode fall inside a band fitted from the reference?} The alternative
asks, per channel, \emph{are the candidate's episodes drawn from the same distribution as the
reference's?} --- a two-sided Mann--Whitney $U$ test, Holm--Bonferroni corrected across each instance's
numeric channels, so this is not $p$-hacking by channel count.

\begin{table}[t]
\caption{Decision-rule sweep on the counterfeit suite: same episodes, same channels, different aggregation. The shipped per-seed rule catches nothing; a distributional rule recovers a third of the suite at no cost in false rejects. The residual at every operating point is the observability bound.}
\label{tab:rules}
\centering\small
\begin{tabular}{@{}lrr@{}}
\toprule
\textbf{Decision rule} & \textbf{Counterfeits rejected} & \textbf{Correct fixes wrongly rejected} \\
\midrule
Shipped: per-seed tolerance bands & 0/11 & 0/5 \\
Distributional, $\alpha=0.001$ & 1/11 & 0/5 \\
\textbf{Distributional, $\alpha=0.01$} & \textbf{4/11} & \textbf{0/5} \\
Distributional, $\alpha=0.05$ & 7/11 & 1/5 \\
\bottomrule
\end{tabular}
\end{table}

Table~\ref{tab:rules} gives the sweep. At $\alpha = \AlphaBest{}$ the distributional rule rejects
\NRuleBestCaught{}/\NRuleCf{} counterfeits while wrongly rejecting \NRuleBestFR{}/\NRulePos{} correct
fixes --- a free lunch on this data. Loosening to $\alpha = \AlphaLoose{}$ nearly doubles detection to
\NRuleLooseCaught{}/\NRuleCf{} but starts costing false rejects (\NRuleLooseFR{}/\NRulePos{}), which
is precisely the trade-off curve two-sided measurement exists to expose. So a third of the acceptance
failure is an \emph{aggregation} choice we can fix, at no cost, by asking a distributional question of
data we already collect. We report this as measured evidence for the next battery version, not as a
change to the shipped oracle: the rule is frozen for the results in this paper, and changing it is a
version bump with a republished table.

\paragraph{The residual is an observability bound.} Between \NRuleBestResidual{} and
\NRuleLooseResidual{} of \NRuleCf{} counterfeits survive every rule we swept, and that residual is the
result worth generalising. Let $o = \pi(\sigma)$ be the observation the oracle grades, for system state
$\sigma$. A counterfeit and the reference are separable by \emph{any} predicate over $o$ only if
$\pi(\sigma_{cf}) \neq \pi(\sigma_{ref})$. Where $\pi$ collapses the difference, no criterion, no
additional episode and no additional mission recovers it, because the recoverable information is gone
before the oracle is consulted. The quantifier covers semantics as well as thresholds: a
falsification-style robustness margin \cite{fainekos2009robustness} is another functional of the
same $o$, so continuous scoring moves an oracle within this bound, not past it. The surviving
counterfeits are equivalent mutants \emph{with respect
to the observation channel} --- a far weaker and more easily satisfied condition than the semantic
equivalence studied in mutation testing \cite{papadakis2015trivial}. The novelty is not that
channel-equivalent variants exist but that here the projection, not program semantics, is the binding
constraint, with a measurable cost: the gap between what the best rule catches and what an omniscient
oracle would. A unit-test benchmark has no analogous gap; its harness asserts over program state and
$\pi$ is near-identity.

Concretely: in one AMCL instance a callback reassigns a divisor without the startup guard, so setting
it to zero divides by zero; the reference clamps it. The read-only-parameter counterfeit validates
identically, warns, returns success --- and assigns to a new sibling member written once and read
nowhere. On the fault mission the two produce \emph{identical} runtime state; on the control mission
the counterfeit does ignore a legitimate stimulus, but resampling every update instead of every fourth
does not move the robot differently.

\paragraph{Loud and silent counterfeits.} Mining logs we record but do not grade,
only \NControlAudible{} of \NCounterfeit{} counterfeits are audible on the control mission, and they
are exactly the over-broad-rejection pair: rejecting a \emph{legal} configuration is loud; the rest
accept, report success, and discard silently. Consistently, all \NSignalChannels{} channels that
ever carried a distributional signal are trajectory summaries. The state separating the silent
counterfeits is internal --- which parameter the node actually used --- and we do not export it.

\paragraph{Log-diffing is a trap, not the fix.} A naive log-diff appears to flag six of
\NCounterfeit{}, but almost all of that is an artifact of our own authoring --- both arms warn,
differing only in wording we chose --- and a counterfeit copying the reference's string verbatim
erases the signal entirely.

\section{Experiments}
\label{sec:eval}

\subsection{Setup}

We evaluate three frontier models --- Claude Sonnet~5, Opus~5 and Fable~5 --- over the full
\NAdmittedEval{}-instance corpus, giving \NCandidates{} model--instance cells.

\paragraph{Attempts.} Each model gets exactly \NAttempts{} attempt per instance: no retries, no
best-of-$N$, no reflection on harness feedback, no iterative repair loop; the patch is scored as
submitted. Rates are therefore single-sample
pass@1, a lower bound on what iteration would achieve, and that choice is the dominant source of
variance here; we quantify it below. Each candidate is
then compiled into the stack and evaluated over \NSeeds{} replicate episodes per mission (unseeded;
see \S\ref{sec:limits}), with the criterion verdict aggregated across replicates as in
\S\ref{sec:bench}.

\paragraph{Prompt.} Reproduced verbatim in Appendix~\ref{app:prompt}. It is machine-generated,
identical in structure across all \NAdmittedEval{} instances, and contains no issue prose, no
upstream reference and no hint at the fix; because prompt content plausibly affects the numbers, we
state it in full rather than describing it.

\paragraph{Excluded rows.} \NDroppedPreContract{} rows produced under a retired patch-application path
are excluded on provenance grounds rather than aged down. We measured what that costs: of the
\NContractCompared{} instances evaluated under both paths, \NContractAgree{} give an identical
outcome label, and the exclusion buys one corrected verdict --- the sole cell that ever occupied
\texttt{fixed\_but\_regressed}.

\subsection{Results}

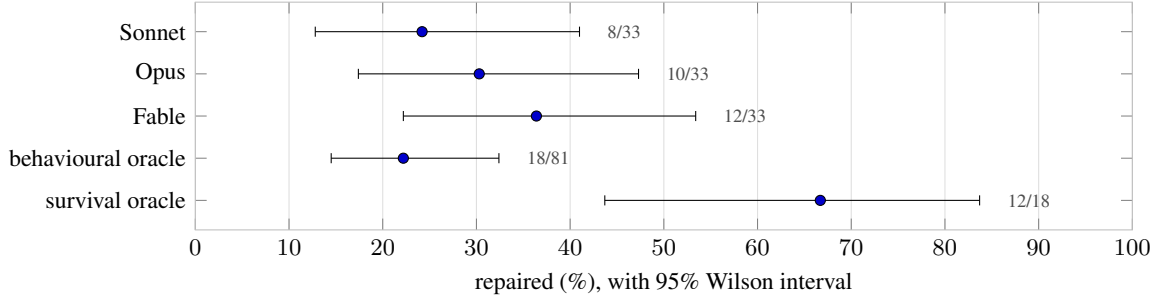
\begin{figure}[t]
\centering
\begin{tikzpicture}
\begin{axis}[width=\linewidth,height=4.6cm,xmin=0,xmax=100,
  xlabel={repaired (\%), with 95\% Wilson interval},
  ytick=data,yticklabels={Sonnet,Opus,Fable,behavioural oracle,survival oracle},
  ymin=-0.7,ymax=4.7,y dir=reverse,
  tick label style={font=\small},label style={font=\small},
  xmajorgrids,grid style={gray!25},axis line style={gray!50}]
\addplot+[only marks,mark=*,mark size=1.9pt,black,
  error bars/.cd,x dir=both,x explicit,error bar style={black}]
  coordinates {
    (24.2,0) +- (16.8,0) -= (11.4,0)
    (30.3,1) +- (17.0,0) -= (12.9,0)
    (36.4,2) +- (17.0,0) -= (14.2,0)
    (22.2,3) +- (10.2,0) -= (7.7,0)
    (66.7,4) +- (17.0,0) -= (23.0,0)
  };
  \node[anchor=west,font=\scriptsize,gray!60!black] at (axis cs:43.0,0) {8/33};
  \node[anchor=west,font=\scriptsize,gray!60!black] at (axis cs:49.3,1) {10/33};
  \node[anchor=west,font=\scriptsize,gray!60!black] at (axis cs:55.4,2) {12/33};
  \node[anchor=west,font=\scriptsize,gray!60!black] at (axis cs:34.4,3) {18/81};
  \node[anchor=west,font=\scriptsize,gray!60!black] at (axis cs:85.7,4) {12/18};
\end{axis}
\end{tikzpicture}
\caption{Repair rate with 95\% Wilson intervals, 1 attempt per instance. The three model intervals overlap heavily (every pairwise exact McNemar $p \geq 0.21$), so this corpus cannot rank them. The bottom pair pools all 99 cells by oracle shape and is the one separation the corpus does support (instance-level $p = 0.0027$; pooled-cell $p = 0.0005$ is anticonservative --- cells cluster by instance).}
\label{fig:ci}
\end{figure}

\begin{table}[t]
\caption{Outcome taxonomy per model over the \NAdmittedEval{}-instance corpus, \NAttempts{} attempt
per cell. \texttt{harness\_error} is reported but excluded from rates; it is zero for all three
models. Intervals on the repaired column are in Figure~\ref{fig:ci}. \texttt{empty\_patch},
\texttt{no\_submission} and \texttt{invalid\_submission} are omitted: every model submitted a
well-formed patch for every instance.}
\label{tab:models}
\centering
\small
\begin{tabular}{@{}lrrrrrrr@{}}
\toprule
\textbf{Model} & \textbf{Scored} & \textbf{Repaired} & \textbf{\%} & \textbf{Fixed but} &
\textbf{Failed and} & \textbf{Ran and} & \textbf{Build /} \\
 & & & & \textbf{regressed} & \textbf{regressed} & \textbf{failed} & \textbf{unapplyable} \\
\midrule
Sonnet & \NSonnetScored{} & \NSonnetRepaired{} & \PctSonnetRepaired{} &
\NSonnetFixedButRegressed{} & \NSonnetFailedAndRegressed{} & \NSonnetRanAndFailed{} &
\NSonnetBuildFailure{} / \NSonnetUnapplyable{} \\
Opus & \NOpusScored{} & \NOpusRepaired{} & \PctOpusRepaired{} & \NOpusFixedButRegressed{} &
\NOpusFailedAndRegressed{} & \NOpusRanAndFailed{} & \NOpusBuildFailure{} / \NOpusUnapplyable{} \\
Fable & \NFableScored{} & \NFableRepaired{} & \PctFableRepaired{} & \NFableFixedButRegressed{} &
\NFableFailedAndRegressed{} & \NFableRanAndFailed{} & \NFableBuildFailure{} / \NFableUnapplyable{} \\
\bottomrule
\end{tabular}
\end{table}

\paragraph{Headline.} Table~\ref{tab:models} gives the full outcome taxonomy. Fable repairs
\NFableRepaired{}/\NFableScored{} (\PctFableRepaired\%, Wilson 95\% CI [\CIFableLo\%, \CIFableHi\%]),
Opus \NOpusRepaired{}/\NOpusScored{} (\PctOpusRepaired\%, [\CIOpusLo\%, \CIOpusHi\%]), and Sonnet
\NSonnetRepaired{}/\NSonnetScored{} (\PctSonnetRepaired\%, [\CISonnetLo\%, \CISonnetHi\%]). We report
the ordering as measured, not as a capability ranking: all three intervals overlap, every pairwise
exact McNemar test (paired: every model scores the same instances)
returns $p \geq \PModelMin{}$, and at $n = \NAdmittedEval{}$ single-sample
instances the corpus size, not the model, binds what a between-model comparison can support. The
result the corpus \emph{does} support
is the level: with one attempt and no repository context, roughly three quarters of these defects go
unrepaired by every model tested.

\paragraph{These rates are bounds, in both directions.} \S\ref{sec:bound} establishes that the shipped oracle
accepts deliberately constructed patches that suppress a failure without repairing the defect, so an
unknown share of the \texttt{repaired} verdicts may be of that kind --- an \emph{upward} bias the
oracle alone cannot bound. The opposite bias is computable (\S\ref{sec:disc}: unanimity over stochastic replicates rejects
correct fixes at rate $1 - q^{\NSeeds{}}$), biasing the rates \emph{downward}. Two considerations narrow the upward share:
the counterfeit authors knew defect and criteria, unlike the solver, and solvers are denied the
instance records and thresholds --- neither is a measurement. The
honest statement is that \PctSonnetRepaired{}--\PctFableRepaired\% is what \emph{this oracle at this
operating point} accepts; we release the accepted patches so a third party can check specific ones
without re-executing the benchmark.

\paragraph{Failure is dominated by wrong fixes, not by refusal.} Every model submitted a well-formed,
applyable patch for every instance, and only one build failure occurs across all \NCandidates{}
cells; the mass sits in \texttt{ran\_and\_failed} --- patches that compiled, ran, and left the robot
failing its mission. Models are not declining these tasks; they are confidently producing plausible
patches that do not fix the defect.

\texttt{fixed\_but\_regressed}, in turn, is zero for all
three models: no model bought an F2P pass by breaking a P2P behavior. The two regressions observed are
\texttt{failed\_and\_regressed} --- patches that were simply wrong. We had pre-registered the opposite
expectation, so this is an informative negative: the functionality-deleting shape the P2P battery
exists to catch (reproduced deliberately in \S\ref{sec:bound}, and unprompted by a model on 5 of 5
episodes under a weaker oracle, \S\ref{sec:oracle}) is not what these models do spontaneously when
asked for a minimal fix: the anti-gaming guard did not need to fire.

\paragraph{Difficulty tracks the oracle, not the domain.} Pooling all \NCandidates{} cells, instances
whose battery is dominated by process survival --- did the stack stay alive --- are repaired
\NCrashRepaired{}/\NCrashCells{} (\PctCrashRepaired\%), while instances graded on what the robot
actually did are repaired \NBehavRepaired{}/\NBehavCells{} (\PctBehavRepaired\%). Pooled cells give
Fisher $p = \PCrashVsBehav{}$, an anticonservative figure --- cells cluster three-per-instance, with
\NInstUnanimous{} of \NAdmittedEval{} instances unanimous across models --- so we cite the
instance-level test instead: at least one model repairs \NCrashInstRepaired{}/\NCrashInst{}
crash-survival against \NBehavInstRepaired{}/\NBehavInst{} behavioral instances, $p = \PCrashVsBehavInst{}$. Crash-survival
oracles resemble unit tests --- near-binary, near-deterministic, satisfied by any crash-removing
patch --- so models doing three times better there is consistent with behavioral grading, not ROS
or C++ unfamiliarity, making the rest of the corpus hard --- though the contrast crosses oracle
families; the survival-only re-grade that would separate the two has not been run.

Pooling instead by architectural layer:
runtime configuration \NLTwoRepaired{}/\NLTwoCells{} cells repaired, perception \NLFourRepaired{}/\NLFourCells, planning \NLThreeRepaired{}/\NLThreeCells, and \emph{zero} of \NLOneCells{} and \NLSixCells{} respectively in
the two layers whose defects sit inside algorithmic or behavior-tree engine logic --- by every
model: small strata we read no rate off, but the pattern identifies where corpus growth is most
valuable.

\paragraph{Verdict instability is the noise floor.} On an earlier 15-instance snapshot, the same
model run twice under two elicitation modes (returning a unified diff vs.\ the complete post-fix
file) agreed on the repaired/not-repaired label for only 9 of
15 instances, so roughly 40\% of per-instance verdicts change
between draws. This conflates sampling variance with an elicitation-mode effect, so it establishes
that verdicts are unstable rather than an exact rate --- which suffices: any between-model difference
smaller than this is not interpretable.

\paragraph{Self-authored instances run easier --- a selection effect, not circularity.}
\NNoPublicFix{} instances carry a reference fix we authored because none exists upstream --- the
synthetic-instance concern \cite{liang2026illusion,huang2026deepswe} pointed at ourselves. Pooled
over the three models they repair at \NSelfAuthoredRepaired{}/\NSelfAuthoredTotal{} cells against
\NUpstreamRepaired{}/\NUpstreamTotal{}; at the instance level, \NSelfAuthoredInstRepaired{}/\NSelfAuthoredInst{}
against \NUpstreamInstRepaired{}/\NUpstreamInst{} ($p = \PProvInst{}$). Authorship circularity is
ruled out by the interaction: \SelfGoldAuthor{}, the model that wrote every self-authored gold, shows
the \emph{smallest} self-vs-upstream advantage (+\DeltaSonnetProv{}\,pp, against +\DeltaOpusProv{}
and +\DeltaFableProv{}\,pp for the two non-authors) --- the opposite of what self-evaluation bias
predicts. The likelier account is selection in corpus construction: the stratum admits only defects
tractable enough for us to author and validate a fix (Appendix~\ref{app:strata}).

\paragraph{Memorization, checked rather than asserted.} Contribution~4's contamination claim is
checkable: instances whose upstream fix merged at or before the pin cut --- plausibly in every
model's training window --- repair at \NExposedRepaired{}/\NExposedCells{} cells, the same rate as
the unexposed remainder (\NUnexposedRepaired{}/\NUnexposedCells{}; $p = \PExposure{}$): exposure
explains none of the variance, and \emph{never having been public} (\NNoPublicFix{} instances)
remains the stronger guarantee.

\section{Discussion}
\label{sec:disc}

\paragraph{What the level means.} Three frontier models, each given a real defect, the exact source
files containing it, and a description of how the robot failed, repair between a quarter and a third
on one attempt. The corpus is not adversarial --- reference fixes are a median of \PatchMedianLines{}
lines and the solver is told which files to read --- so the gap is not repository-scale search but
reasoning from observed behavior back to the control code that produced it: the capability robot
software repair requires.

\paragraph{What transfers from test-graded benchmarks, and what does not.} Instance anatomy, the
F2P/P2P split and the withheld oracle transfer directly. Three things do not: determinism (verdicts
must aggregate over replicates);
non-regression as a free by-product (the cheapest passing patch often disables behavior, so P2P must
be built deliberately); and the standard remedy for an inadequate oracle (\S\ref{sec:bound}).

\paragraph{Are \NSeeds{} replicates enough? No, and the shortfall is computable.} A candidate's
criterion verdict is taken by \emph{unanimity} over its \NSeeds{} episodes, so a correct patch with
per-episode pass probability $q$ is accepted with probability $q^{\NSeeds{}}$ --- at $q = 0.9$,
$0.73$: roughly a quarter of correct fixes on a mildly stochastic instance are rejected by run-to-run
noise alone, a systematic downward bias and a second reason to read our rates as bounds. (The
\emph{target} sets are not exposed this way: F2P and P2P derive from \NBuggyEpisodes{} cached buggy
episodes, not three.) We chose \NSeeds{} because each episode is a full Gazebo rollout and the
campaign is \NCandidates{} cells deep; repeated sampling on both axes is the first use for more
compute.

\paragraph{Are \NAdmittedEval{} instances enough to rank models? Also no, computably.} Model
pairs separate only on \emph{discordant} instances --- those exactly one of the two repairs ---
and the widest pair has \NDiscordantWidest{}: a \PctModelGapWidest{}pp gap on which the exact
McNemar test has \PctPowerAtN\% power. At the observed discordance rate, 80\% power at
$\alpha = 0.05$ takes \NPowerWidestGap{} instances at that gap and \NPowerHalfGap{} at half of it
--- required $n$ grows roughly quadratically as gaps narrow. Figure~\ref{fig:ci}'s overlap is
thus absence of evidence, not evidence of parity; the corpus is sized for the level, not the
ordering.

\paragraph{Aggregation first, then instrumentation.} For any benchmark graded by what a system does
rather than what it computes, \S\ref{sec:bound} separates two design variables that are easy to
conflate. The first is the \emph{aggregation rule} over observations already collected, and it is
cheap: switching from per-seed tolerance bands to a distributional test recovered
\NRuleBestCaught{}/\NRuleCf{} counterfeits at zero cost in false rejects; a benchmark reporting a
single-episode pass/fail per criterion is leaving detection on the table. The
second is the \emph{observation channel} $\pi$, where the residual lives: counterfeits that survive
every rule do so because the distinguishing state never reaches the channel, not because criteria are
lax. Closing that gap means widening $\pi$ --- exporting planner-internal state and controller
command sequences as first-class graded telemetry. We would spend the next increment of effort on
aggregation (measurably free), then instrumentation, ahead of corpus growth.

\section{Limitations}
\label{sec:limits}

The corpus is concentrated in one navigation stack; extending to manipulation and a second embodiment
is the main scaling axis. Grading is simulation-only; that most real navigation bugs can be revealed
in low-fidelity simulation \cite{sotiropoulos2017navigation} supports the choice, but verdict
transfer to hardware is unmeasured. Reported rates are single-sample pass@1 on a $\approx$40\%
verdict-flip floor.
Replicates in the reported campaign are labelled, not controlled --- the exported per-episode seed
was consumed by no simulation component, so they differ only through uncontrolled scheduler and
physics-timing jitter; seeded spawn-pose perturbation has since landed in the harness, and every
number here predates it.
Most importantly, the oracle demonstrably accepts constructed counterfeits, so reported repair rates
are bounds, not counts of correct fixes --- biased upward by what the oracle cannot
distinguish, downward by that same replicate noise (\S\ref{sec:disc}). The
counterfeit suite covers \NCounterfeit{} of the \NAdmittedCorpus{} instances (the corpus at audit
time), so the acceptance result is measured on a third of what we release. The observability bound is
measured on our channel and fault-operator set --- a different author's suite might fare differently
--- and the oracle-calibration figures rest on single-digit denominators (1 in 6, 2 in 8). Full
third-party re-execution awaits a pinned multi-gigabyte image; the released
counterfeit patches, batteries and cached telemetry make \S\ref{sec:oracle}--\S\ref{sec:bound}
checkable without it.

\section{Conclusion}

Robot software repair has been unmeasurable for want of an oracle to inherit.
RoboBugBench shows one can be constructed --- \NAdmittedCorpus{} real defects in a pinned ROS~2 stack,
graded by compiling a candidate patch and running the robot --- and is informatively
difficult: frontier models repair \PctSonnetRepaired{}--\PctFableRepaired\% on one attempt, with
difficulty concentrated where the oracle stops resembling a unit test: \PctCrashRepaired\% on
process-survival against \PctBehavRepaired\% on behavioral instances (instance-level
$p = \PCrashVsBehavInst{}$).

The oracle work also taught us how to build such benchmarks: audit both error directions, fix
aggregation before criteria, and read the rates as bounds.

\bibliographystyle{plain}
\bibliography{refs}

\appendix

\section{The self-authored versus upstream stratification}
\label{app:strata}

\NNoPublicFix{} of the \NAdmittedCorpus{} instances carry a reference fix we authored because no
upstream fix exists. If those were systematically easier, our headline rates would be flattered ---
and they are easier. Pooled over the three models, self-authored instances repair at
\NSelfAuthoredRepaired{}/\NSelfAuthoredTotal{} cells (\PctSelfAuthoredCells\%) against
\NUpstreamRepaired{}/\NUpstreamTotal{} (\PctUpstreamCells\%) on upstream-backed ones. Because cells
cluster three-per-instance, the primary test collapses to instances: at least one model repairs
\NSelfAuthoredInstRepaired{}/\NSelfAuthoredInst{} self-authored against
\NUpstreamInstRepaired{}/\NUpstreamInst{} upstream instances, two-sided Fisher exact
$p = \PProvInst{}$ (the anticonservative pooled-cell test gives $p = \PProvCells{}$; we cite the
instance-level figure, consistently with \S\ref{sec:eval}).

The explanation matters more than the gap. The damaging reading would be authorship circularity:
every self-authored reference fix was written with \SelfGoldAuthor{} (recorded per instance as
\texttt{authored\_by} in the released metadata), so the stratum could conceivably favor patches in
its author's style. Table~\ref{tab:provenance} rules that reading out: \SelfGoldAuthor{}, the gold
author, shows the \emph{smallest} self-vs-upstream advantage of the three models, and the two
non-authors the largest --- the opposite interaction from what a self-evaluation confound predicts.
The grading oracle itself is execution-based and model-free throughout, so stylistic similarity of
the reference patch was the only channel circularity had, and the interaction shows it does not
carry.

\begin{table}[h]
\caption{Per-model repair rate by reference-fix provenance, one attempt per cell. The model that
authored every self-authored gold shows the smallest advantage on that stratum.}
\label{tab:provenance}
\centering\small
\begin{tabular}{@{}lrrr@{}}
\toprule
\textbf{Model} & \textbf{Self-authored} & \textbf{Upstream} & \textbf{$\Delta$ (pp)} \\
\midrule
Sonnet (gold author) & \NSonnetSelfRepaired{}/\NSelfAuthoredInst{} (\PctSonnetSelf\%) &
\NSonnetUpstreamRepaired{}/\NUpstreamInst{} (\PctSonnetUpstream\%) & $+\DeltaSonnetProv{}$ \\
Opus & \NOpusSelfRepaired{}/\NSelfAuthoredInst{} (\PctOpusSelf\%) &
\NOpusUpstreamRepaired{}/\NUpstreamInst{} (\PctOpusUpstream\%) & $+\DeltaOpusProv{}$ \\
Fable & \NFableSelfRepaired{}/\NSelfAuthoredInst{} (\PctFableSelf\%) &
\NFableUpstreamRepaired{}/\NUpstreamInst{} (\PctFableUpstream\%) & $+\DeltaFableProv{}$ \\
\bottomrule
\end{tabular}
\end{table}

We therefore read the gap as a selection effect of corpus construction rather than a scoring
artifact: an instance enters the self-authored stratum exactly when no upstream fix exists \emph{and}
the defect was tractable enough for this project to author and validate a reference fix, which
plausibly selects for simpler mechanisms independently of which model attempts the repair. The
practical consequence is stratified reporting: both strata are labelled per instance in the released
metadata, and solver comparisons on this benchmark should report the split alongside the pooled rate.
With \NAdmittedEval{} instances and single-attempt scoring on the verdict-instability floor of
\S\ref{sec:eval}, the effect size should be treated as provisional; its direction and the authorship
interaction are the robust parts.

\section{Oracle validation: the failure modes we found on our own harness}
\label{app:oracle}

Each of the following was measured on our own harness, most by probes registered before the result
existed.

\paragraph{A liveness oracle is gameable, as predicted in advance.} Ten of an early fifteen instances
graded only \emph{no collision}, which a stationary robot satisfies. We registered the prediction that
a patch suppressing the crash while leaving the defect intact would score \texttt{repaired}, then ran
it; it did, one cheat scoring \texttt{repaired} having travelled $1.2\times10^{-18}$\,m --- and a model
reproduced the same strategy unprompted on 5 of 5 episodes. This is the origin of the pass-to-pass
battery.

\paragraph{A validation check can be incapable of failing.} An adversarial review found the battery's
central validation check mathematically unable to fail. Freezing the battery before that review would
have frozen a check that could never fire.

\paragraph{An instrument can fail silently and pass its own gates.} A change severed the telemetry
bridge; for roughly a day every episode recorded uniformly-zero telemetry while navigation proceeded
normally. It passed two verification gates because both graded log-derived criteria and never read a
telemetry number. The harness now asserts on instrument variance before it asserts on instrument
values.

\paragraph{Upstream tests do not rescue us.} Of the PR-backed instances checked, only one ships a test
at all, and that test passes on both the buggy and the fixed build --- the test-scarcity premise of
\S\ref{sec:related} confirmed on our own corpus rather than imported.

\section{The solver prompt, verbatim}
\label{app:prompt}

Every model saw exactly the JSON object below and nothing else. It is emitted by the harness, is identical in structure across all 33 instances, and is reproduced here from the on-disk files the campaign ran against.

\paragraph{Task statement (identical for all 33 instances).}
\begin{quote}\small\itshape
You are debugging a REAL, live ROS 2 Jazzy + Nav2 navigation stack bug (not a synthetic injection) running in Gazebo. Below is the observed failure and the current (buggy) artifact --- the actual unmodified pinned source. Propose a MINIMAL fix that makes the robot reach its goal without collision (or, for a crash-survival mission, keeps the stack alive without a process death or lifecycle timeout), changing nothing else.
\end{quote}

\paragraph{Multi-file addendum.} Appended verbatim to the task statement for the 6 instances whose reference fix spans more than one file (with \texttt{N} the file count):
\begin{quote}\small\itshape
NOTE: N related source files are shown below (see `artifacts`), because the defect may span more than one of them --- e.g. a member or method that must be declared in a header before the implementation file can use it. Change as few of them as the fix genuinely requires; a patch that uses a symbol it never declares will not compile.
\end{quote}

\paragraph{Observed failure.} A machine-generated one-line summary of the buggy run's graded criteria --- pass/fail per criterion plus two scalars. No prose diagnosis, no root cause, no hint at the fix. Example (\texttt{rbb\_nav2\_amcl\_R-5734}):
\begin{quote}\small\ttfamily
The stack FAILS its mission. Criteria: no\_collision: PASS, no\_process\_death: FAIL. Final dist\_to\_goal=2.87 m, collision=False.
\end{quote}

\paragraph{Source.} The complete current text of every file the reference fix touches (median 1 file, max 4; median 18,196 source characters, max 60,091), each labelled with its repository-relative path. Nothing else from the repository is shown.

\paragraph{Answer format (single-file, 27 instances).}
\begin{quote}\small\itshape
Return a unified diff (git apply -p1 compatible) against the shown file, i.e. '--- a/<path>' / '+++ b/<path>' with @@ hunks. Only the minimal fix.
\end{quote}

\paragraph{Answer format (multi-file, 6 instances).}
\begin{quote}\small\itshape
Return a unified diff (git apply -p1 compatible). You may edit ONE or SEVERAL of the shown files. Precede each file's hunks with its own '--- a/<path>' / '+++ b/<path>' header using the EXACT paths shown above, so a multi-file edit can be applied. Only the minimal fix.
\end{quote}

\paragraph{What the prompt withholds.} No issue title or body, no commit message, no upstream link or PR number, no test, no repository history, no file list beyond the files shown, no stack trace, no telemetry beyond the two scalars above, and no statement of which criterion the grader will treat as fail-to-pass. Models were additionally run under an integrity contract forbidding them the instance records, the manifest, the benchmark documentation, the git history of the vendored source, and the web.

\section{The full corpus, by taxonomy}
\label{app:corpus}
\small
\begin{longtable}{@{}llllrl@{}}
\toprule
\textbf{Layer} & \textbf{Mechanism} & \textbf{Oracle shape} & \textbf{Mode} & \textbf{Lines} & \textbf{Fix merged} \\
\midrule
\endhead
L1\_algorithm & input\_validation & behavioral & latent & 6 & 2026-04-29 \\
L1\_algorithm & input\_validation & behavioral & latent & 8 & \emph{no public fix} \\
L2\_runtime\_config & behavioral\_degradation & behavioral & latent & 13 & \emph{no public fix} \\
L2\_runtime\_config & dynamic\_reconfigure & behavioral\_velocity & latent & 4 & 2025-11-26 \\
L2\_runtime\_config & dynamic\_reconfigure & crash\_survival & latent & 7 & 2026-04-XX \\
L2\_runtime\_config & dynamic\_reconfigure & crash\_survival & latent & 12 & 2026-04-XX \\
L2\_runtime\_config & dynamic\_reconfigure & behavioral\_abort & latent & 43 & \emph{no public fix} \\
L2\_runtime\_config & dynamic\_reconfigure & crash\_survival & latent & 15 & \emph{no public fix} \\
L2\_runtime\_config & dynamic\_reconfigure & crash\_survival & latent & 13 & \emph{no public fix} \\
L2\_runtime\_config & index\_based\_progress\_tracking & behavioral & latent & 46 & \emph{no public fix} \\
L2\_runtime\_config & input\_validation & behavioral & latent & 6 & 2026-06-26$^{\dagger}$ \\
L2\_runtime\_config & input\_validation & behavioral & latent & 54 & 2026-06-30$^{\dagger}$ \\
L2\_runtime\_config & input\_validation & behavioral & latent & 15 & 2026-07-07$^{\dagger}$ \\
L2\_runtime\_config & input\_validation & behavioral & latent & 9 & 2026-08-14$^{\dagger}$ \\
L2\_runtime\_config & input\_validation & action\_stall & latent & 38 & \emph{no public fix} \\
L2\_runtime\_config & input\_validation & planning\_failure & latent & 19 & \emph{no public fix} \\
L2\_runtime\_config & input\_validation & behavioral\_goal\_abort & latent & 14 & \emph{no public fix} \\
L2\_runtime\_config & iterator\_underflow & behavioral & latent & 31 & 2026-08-10$^{\dagger}$ \\
L2\_runtime\_config & lifecycle\_reconfigure\_not\_idempotent & behavioral & resurrected & 15 & 2023-05-02 \\
L2\_runtime\_config & rotation\_composition\_order & behavioral & latent & 4 & 2026-04-09 \\
L3\_planning & dynamic\_reconfigure & behavioral & latent & 12 & 2026-02-20 \\
L3\_planning & input\_validation & behavioral & latent & 31 & 2025-08-04 \\
L3\_planning & input\_validation & behavioral & latent & 15 & 2025-08-04 \\
L3\_planning & input\_validation & crash\_survival & latent & 22 & \emph{no public fix} \\
L3\_planning & memory\_safety & behavioral & self\_fixed & 15 & \emph{no public fix} \\
L3\_planning & memory\_safety & crash\_survival & latent & 30 & \emph{no public fix} \\
L4\_perception & input\_validation & behavioral & latent & 8 & 2026-07-02$^{\dagger}$ \\
L4\_perception & input\_validation & behavioral & latent & 11 & 2026-06-21$^{\dagger}$ \\
L4\_perception & input\_validation & behavioral & latent & 19 & \emph{no public fix} \\
L4\_perception & memory\_safety & behavioral & resurrected & 17 & 2022-08-24 \\
L4\_perception & sensor\_domain\_conflation & behavioral & self\_fixed & 31 & \emph{no public fix} \\
L6\_bt\_engine & input\_validation & behavioral\_abort & latent & 55 & 2026-04-06 \\
L6\_bt\_engine & input\_validation & behavioral\_position & latent & 31 & 2026-07-22$^{\dagger}$ \\
\bottomrule
\caption{All 33 admitted instances ordered by layer and mechanism. $^{\dagger}$ marks the 8 upstream fixes merged strictly after the 2026-04-29 pin cut, which no model trained before that date can have seen. \emph{no public fix} marks the 14 instances whose reference fix exists in no public record.}
\label{tab:appendix-corpus}
\end{longtable}

\end{document}